\documentclass[sigconf]{aamas}

\usepackage{balance} 
\usepackage{enumitem}
\usepackage{bbm}
\usepackage{booktabs}
\usepackage{array}
\usepackage{adjustbox}
\newcommand{\vgcbench}{\textsc{VGC-Bench}}

\setcopyright{ifaamas}
\acmConference[AAMAS '26]{Proc.\@ of the 25th International Conference
on Autonomous Agents and Multiagent Systems (AAMAS 2026)}{May 25 -- 29, 2026}
{Paphos, Cyprus}{C.~Amato, L.~Dennis, V.~Mascardi, J.~Thangarajah (eds.)}
\copyrightyear{2026}
\acmYear{2026}
\acmDOI{}
\acmPrice{}
\acmISBN{}

\acmSubmissionID{1262}

\title[AAMAS-2026 Formatting Instructions]{VGC-Bench: Towards Mastering Diverse Team Strategies \\ in Competitive Pok\'emon}

\author{Cameron L.~Angliss}
\affiliation{
  \institution{University of Texas at Austin}
  \city{Austin, TX}
  \country{United States}}
\email{cangliss@utexas.edu}

\author{Jiaxun Cui}
\authornote{Equal contribution}
\affiliation{
  \institution{University of Texas at Austin}
  \city{Austin, TX}
  \country{United States}}
\email{cuijiaxun@utexas.edu}

\author{Jiaheng Hu}
\authornotemark[1]
\affiliation{
  \institution{University of Texas at Austin}
  \city{Austin, TX}
  \country{United States}}
\email{jiahengh@utexas.edu}

\author{Arrasy Rahman}
\affiliation{
  \institution{University of Texas at Austin}
  \city{Austin, TX}
  \country{United States}}
\email{arrasy@cs.utexas.edu}

\author{Peter Stone}
\authornote{Sony AI}
\affiliation{
  \institution{University of Texas at Austin}
  \city{Austin, TX}
  \country{United States}}
\email{pstone@cs.utexas.edu}

\begin{abstract}
Developing AI agents that can robustly adapt to varying strategic landscapes without retraining is a central challenge in multi‑agent learning. Pokémon Video Game Championships (VGC) is a domain with a vast space of approximately $10^{139}$ team configurations, far larger than those of other games such as Chess, Go, Poker, StarCraft, or Dota. The combinatorial nature of team building in Pokémon VGC causes optimal strategies to vary substantially depending on both the controlled team and the opponent's team, making generalization uniquely challenging. To advance research on this problem, we introduce \vgcbench: a benchmark that provides critical infrastructure, standardizes evaluation protocols, and supplies a human-play dataset of over 700,000 battle logs and a range of baseline agents based on heuristics, large language models, behavior cloning, and multi-agent reinforcement learning with empirical game-theoretic methods such as self-play, fictitious play, and double oracle. In the restricted setting where an agent is trained and evaluated in a mirror match with a single team configuration, our methods can win against a professional VGC competitor. We repeat this training and evaluation with progressively larger team sets and find that as the number of teams increases, the best-performing algorithm in the single-team setting has worse performance and is more exploitable, but has improved generalization to unseen teams. Our code and dataset are open-sourced at \url{https://github.com/cameronangliss/vgc-bench} and \url{https://huggingface.co/datasets/cameronangliss/vgc-battle-logs}.

\end{abstract}


\newcommand{\BibTeX}{\rm B\kern-.05em{\sc i\kern-.025em b}\kern-.08em\TeX}

\begin{document}


\pagestyle{fancy}
\fancyhead{}


\maketitle 


\section{Introduction}

Pokémon is the highest-grossing media franchise in the world, with an estimated total revenue exceeding \$100 billion and a global player base numbering in the millions. In 2008, the Pokémon Company launched the Video Game Championships (VGC), a competitive series of tournaments featuring significant cash prizes and international prestige. The largest tournament to date, EUIC 2025, featured 1,257 competitors. Despite this popularity and competitive depth, to the best of our knowledge, no AI system has yet achieved superhuman performance in competitive Pokémon VGC battles.

Competitive Pokémon presents two tightly-coupled challenges: team building and team usage. Note that we only address team usage in this work; team building is left as an open challenge. We estimate the size of the team configuration space to be on the order of $10^{139}$, vastly exceeding the configuration space of other benchmark games such as Dota or StarCraft. Because of the discrete and combinatorial nature of team-building, optimal strategies can vary dramatically depending on the team compositions of both the player and opponent. As a result, even expert human players struggle to generalize strategies across matchups -- even when their own team remains fixed, as is standard in VGC tournaments, but especially if trying to pilot many different team compositions. Some teams focus on controlling the weather; some focus on controlling the speed of Pokémon on the field; some focus on preventing the opponent's Pokémon from switching out. There is no well-defined maximum number of strategic paradigms with which a team can be built around, and many teams are hybrid, playing to more than one of these powerful strategies at the same time.

This combination of a vast space of team configurations and highly diverse team strategies makes Pokémon an especially valuable testbed for generalization in AI systems. In the past, research has been conducted on Pokémon with heuristic and search-based agents \citep{panumate2016pokemon, lee2017showdown, karten2025pok}, reinforcement learning \citep{huang2019self,simoes2020competitive, wang2024winning, pleines2025pokemon}, team building \citep{simoes2023adversarial}, and predicting the winner at any state \citep{charde2019predicting}. However, most of the prior work focuses exclusively on the Single Battle format, where each player sends out one Pokémon at a time. This setting significantly simplifies the underlying decision space: in contrast, the official Pokémon VGC tournaments use Double Battle format, which involves two active Pokémon per side, leading to a combinatorial explosion in possible actions and interactions and more challenging learning.

\begin{figure*}[t]
    \centering
    \includegraphics[width=0.9\linewidth]{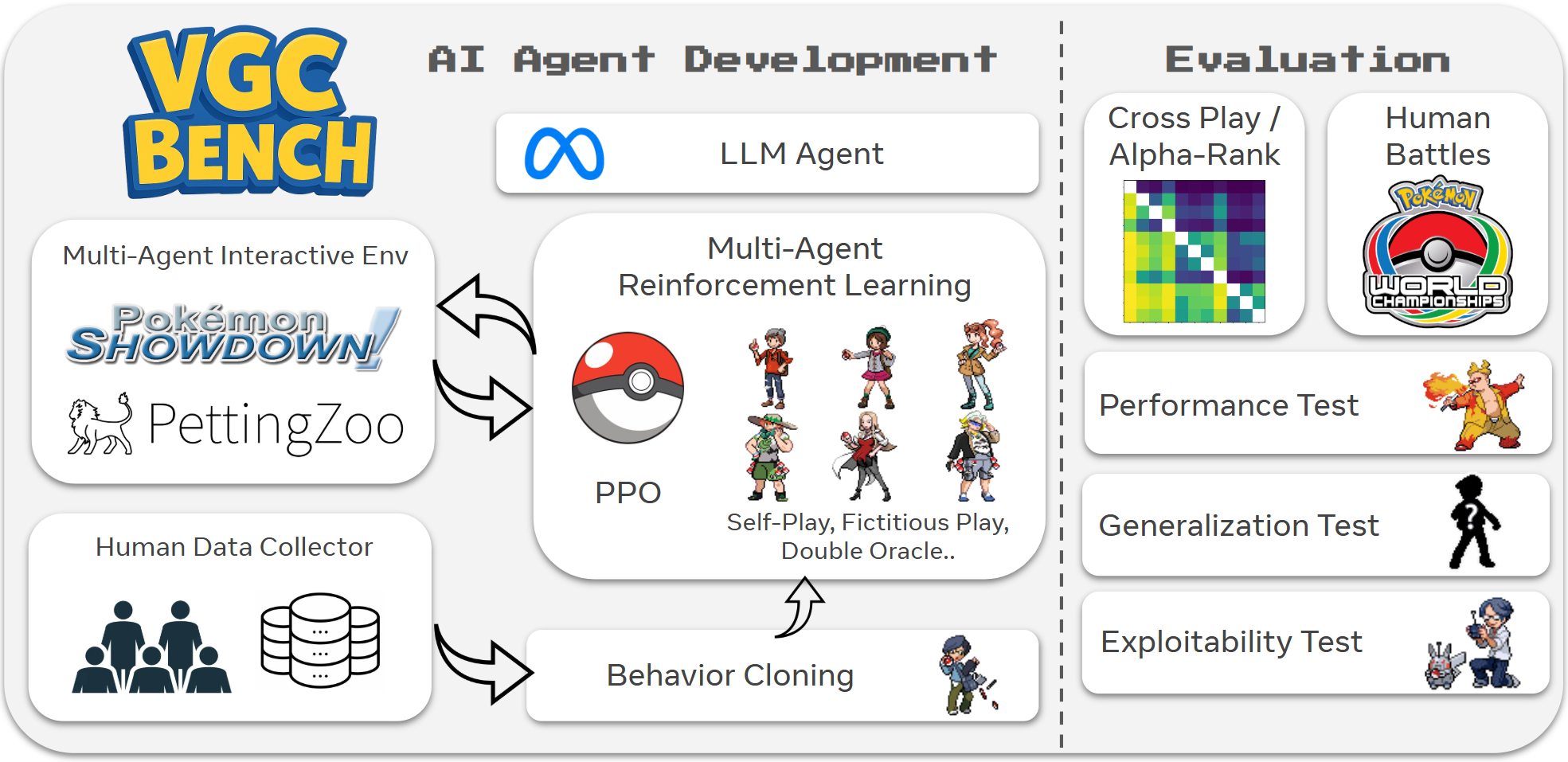}
    \caption{\textit{\small\textbf{\vgcbench} Overview}. \vgcbench{} captures the multi-agent multi-team dynamics with PettingZoo integration, provides human-play datasets and a range of baselines, and standardizes evaluation protocols.}
    \label{fig:flowchart}
    \Description{\vgcbench{} captures the multi-agent multi-team dynamics with PettingZoo integration, provides human-play datasets and a range of baselines, and standardizes evaluation protocols.}
\end{figure*}

In this work, we present \vgcbench, a benchmark designed to evaluate AI generalization in Pokémon VGC. Our contributions include infrastructure for multi-agent learning and human-play data collection, including a dataset of over 700,000 open team sheet VGC battle logs; a suite of competitive baselines, covering multi-agent RL, behavior cloning, using a language model, and heuristic agents; and robust evaluation tools for performance, generalization, exploitability, and human interaction.
In the restricted setting where an agent is trained and evaluated on a single-team configuration, our methods are able to win against a professional VGC competitor. We then extend evaluation to broader multi-task settings involving diverse team configurations. While existing methods can learn competent policies in narrow conditions, they exhibit notable performance degradation as team diversity increases. These findings underscore the need for more generalizable learning algorithms in multi-agent, multi-task environments like VGC.

Furthermore, as part of our work, we contributed significant open-source contributions to \texttt{Poke-env} \citep{poke_env}, a widely used library that has supported prior research in Pokémon AI. Our contributions include full integration with the PettingZoo \citep{terry2021pettingzoo} multi-agent framework, extended support for VGC and doubles formats, and many bugfixes to correct battle tracking and more. We expect that these enhancements will enable and encourage more accessible and rigorous experimentation, further advancing AI research in this domain.

\section{Problem Formulation}

\subsection{Game Mechanics}
Pokémon VGC is a team-based competitive game with stochastic dynamics, simultaneous move selection, and a vast configuration space of possible teams. In the VGC format, each player assembles a team of 6 Pokémon, with each Pokémon configured by customized individual stats, up to 4 moves, a passive ability, a Tera type, and potentially a held item. A Tera type is a gimmick mechanic specific to generation 9 of Pokémon which defines a type (i.e. grass, fire, water) that a Pokémon can overwrite their default types to for the rest of the battle. Players can only terastallize a Pokémon once per battle, and is commonly used defensively to avoid being super-effectively hit by an attack, but can also be used offensively to boost the damage of moves of the same type. The VGC format is entirely set in \textit{double} battles, where players deploy up to 2 Pokémon on the field at a time. Recent tournaments adopt \textit{Open Team Sheets (OTS)}, a modification on the VGC format that reveals nearly all aspects about the opponent’s team -- such as moves, items, abilities, and Tera types -- while leaving the precise underlying stats concealed. Each match begins with a \textit{Team Preview} phase, during which players simultaneously select 4 out of their 6 Pokémon to bring into the battle. The first two chosen are sent out at the start of the match, while the other two remain in reserve and can be switched in as the battle progresses. Once the battle begins, both players issue commands independently and simultaneously each turn, and only after the players have locked in their decisions do the battle mechanics determine the order of events for that turn. Generally, Pokémon switch out before moves are performed, and the execution order of moves is usually determined by the speed stats of the acting Pokémon, with speed ties broken randomly. The objective is to knock out all of the opponent’s Pokémon by reducing their HP to zero before your own team is eliminated.

Several factors make VGC particularly challenging for AI agents (see Appendix \ref{sec:game-analysis} for calculation of values presented here):
\begin{enumerate}[leftmargin=*]
    \item \textbf{Combinatorial Team Configurations}: The team configuration space in Pokémon VGC is large. With hundreds of Pokémon species, moves, items, and abilities, 19 possible Tera types per Pokémon, and a high-dimensional space of possible stat allocations, we estimate the total number of valid team configurations to be approximately $10^{139}$. This combinatorial space far exceeds that of many other strategic games (see Table~\ref{tab:game-comparison} for comparison).
    \item \textbf{Stochastic Battle Mechanics}: Move outcomes in VGC involve randomness in damage and secondary effects (e.g., 10\% chance to paralyze). This results in a large branching factor per turn, which we approximate to be at times as large as $10^{12}$.
    \item \textbf{Partial Observability}: Even with OTS, Pokémon VGC remains a partially observable game as the opponent's stats are not revealed. We estimate that the size of the information set -- the set of all possible states given a partial observation of the game -- is lower-bounded at approximately $10^{58}$.
    \item \textbf{Simultaneous and Multi-Agent Actions}: Pokémon VGC is a simultaneous game that requires up to four Pokémon to make a decision at the same time. This game feature introduces non-stationarity and pose challenges to accurate credit assignment during policy optimization. For example, suppose both of a player's Pokémon target the same opponent Pokémon, and one of them uses a move the target is immune to, but the other knocks the Pokémon out in one hit. In this case, assuming the AI only observes the game state before and after the turn, the AI could never know which Pokémon caused the knock out.
    \item \textbf{Team Preview Strategy and Generalization:} Pokémon VGC features a unique challenge: only four of the six team members are used in each battle. This introduces a mini team-selection problem, where the player must pick the most effective subteam against the opponent’s lineup. This single decision can drastically alter the dynamics of the battle, and with a total of $\binom{6}{2} \cdot \binom{4}{2} = 90$ possible team preview decisions for each player, effective exploration and generalization are especially challenging.
\end{enumerate}

\subsection{Formalization}
We model each Pokémon battle as a two-player zero-sum partially observable stochastic game (POSG) with randomized team configurations. Let $\mathcal{C}$ denote the finite set of legal team configurations and $\mathcal{U}(\mathcal{C})$ denote the uniform distribution over those team configurations. At the start of each episode, players independently and uniformly sample $c_1, c_2 \sim \mathcal{U}(\mathcal{C})$, which instantiates
\[
\mathcal{G}(c_1, c_2) = \langle \mathcal{S}, \{\mathcal{O}_i\}, \{\mathcal{A}_i\}, \mathcal{T}, R, c_1, c_2 \rangle.
\]
Here, $\mathcal{S}$ is the full battle state (HP, status, weather, stat boosts, etc.), $\mathcal{O}_i \subseteq \mathcal{S}$ is player $i$'s observation, $\mathcal{A}_i$ the action set (switching, move selection, targeting, terastallization), $\mathcal{T}(s' \mid s,a_1,a_2)$ the stochastic transition function, and $R(s_T) \in \{ -1, 1\}$ the terminal reward for player~1.

Given policies $\pi_1,\pi_2$ mapping observations to action distributions, the distribution of trajectories initialized with team configurations $c_1, c_2$ can be defined as
\begin{equation}
    \Gamma_{c_1, c_2}(\pi_1, \pi_2) = \prod_{t=0}^{T-1}
    \pi_1(a_1^t \mid o_1^t)
    \pi_2(a_2^t \mid o_2^t)
    \mathcal{T}(s^{t+1} \mid s^t,a_1^t,a_2^t).
    \label{eq:traj_dist}
\end{equation}
The expected return for player 1 over all team draws is
\begin{equation}
    V(\pi_1,\pi_2)
    = \mathbb{E}_{c_1,c_2 \sim \mathcal{U}(\mathcal{C})}
      \left[
      \mathbb{E}_{\tau \sim \Gamma_{c_1, c_2}(\pi_1, \pi_2)}[R(s_T)]
      \right].
    \label{eq:expected_value}
\end{equation}
Since the game is zero-sum, the equilibrium in expectation over configurations is
\begin{equation}
    (\pi_1^{\text{eq}},\pi_2^{\text{eq}})
    = \arg\max_{\pi_1}
      \arg\min_{\pi_2}
      V(\pi_1,\pi_2).
    \label{eq:minimax}
\end{equation}
This formalism captures our objective: to find robust policies that perform well on average against all possible opposing team draws in the configuration space.

\begin{figure}[t]
    \centering
    \includegraphics[width=\columnwidth]{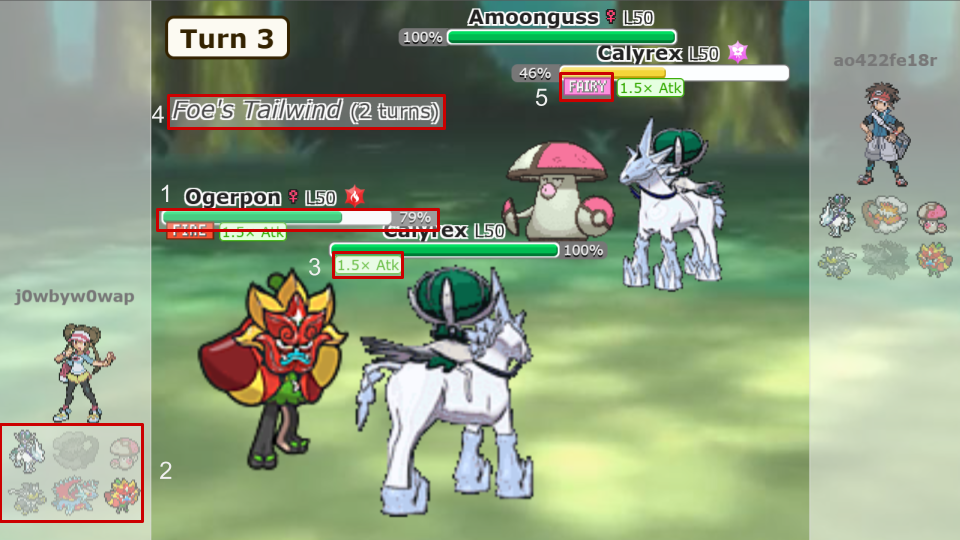}
    \caption[Pokémon Showdown Gameplay]{Pokémon Showdown Gameplay. Closer Pokémon are on the agent's side, and farther Pokémon are on the opponent's side. 1) Pokémon's health bar with percentage of current health remaining. 2) Current status of all party members, with solid colors for revealed, translucent colors for unrevealed, and greyed-out colors for fainted. 3) Effects on Pokémon, including boosts and status conditions. 4) Active side conditions and global fields/weather with a number of turns remaining. 5) Active Tera type being used by Pokémon.}
    \label{fig:gameplay}
    \Description{Pokémon Showdown Gameplay. Closer Pokémon are on the agent's side, and farther Pokémon are on the opponent's side. 1) Pokémon's health bar with percentage of current health remaining. 2) Current status of all party members, with solid colors for revealed, translucent colors for unrevealed, and greyed-out colors for fainted. 3) Effects on Pokémon, including boosts and status conditions. 4) Active side conditions and global fields/weather with a number of turns remaining. 5) Active Tera type being used by Pokémon.}
\end{figure}

\section{Related Work}

\subsection{Progress in Game AI.}
Early AI milestones focused on perfect-information games like Chess and Go, where IBM’s \textit{Deep Blue} and DeepMind’s AlphaGo achieved superhuman performance through massive search and pattern recognition \citep{campbell2002deep, silver2016mastering}. Poker introduced reasoning under partial observability and uncertainty, with Libratus defeating top humans in no-limit Texas Hold’em \citep{brown2017libratus}. Recent work shifted to more complex multi-agent games: AlphaStar reached grandmaster in StarCraft II by population-based reinforcement learning \citep{vinyals2019grandmaster} and OpenAI Five mastered Dota 2 with limited hero drafting \citep{berner2019dota}. In contrast, Pokémon VGC offers a unique challenge that combines partial observability, stochasticity, simultaneous actions, and a combinatorially vast team configuration space larger than any prior benchmarked game (see Table~\ref{tab:game-comparison}).

\begin{table}[h]
  \centering 
  \caption[Configuration Space Comparison]{Configuration space comparison of benchmark games versus Pokémon VGC. Calculations for numbers can be found in section \ref{sec:game-analysis}.}
  \label{tab:game-comparison}
  \begingroup
  \renewcommand{\arraystretch}{1.2}
  \begin{tabular}{@{} lccccc @{}}
    \toprule
    \textbf{Game} & \textbf{Init. Config. Space} \\
    \midrule
    Chess & 2 \\
    Go & 2 \\
    Poker (10 players) & $\binom{52}{2} \binom{50}{2} \cdots \binom{34}{2} \approx 10^{29}$ \\
    StarCraft II & $(3~\texttt{races})^{2~\texttt{players}} * (9~\texttt{maps}) = 81$ \\
    Dota 2 & $\binom{2 * 126}{5}\binom{2 * 121}{5} \approx 10^{20}$ \\
    \textbf{Pokémon VGC} & $\approx (10^{139})^{2~\texttt{players}} = 10^{278}$ \\
    \bottomrule
  \end{tabular}
  \renewcommand{\arraystretch}{1.0}
  \endgroup

\end{table}

\subsection{Existing Pokémon AI}

\subsubsection{Heuristic and Search-Based Agents} Early Pokémon AI work relied on handcrafted rules and search. Lee and Togelius (2017) introduced the \emph{Showdown AI Competition}, a benchmark based on a clone of Pokémon battles \citep{lee2017showdown}. They highlighted that Pokémon battles involve turn-based team combat with partial observability, a challenging setup uncommon in prior AI competitions. Huang and Lee (2019) note that before DRL, most agents used expectimax or minimax search with learned evaluations. Panumate and Iida (2016) built a simplified Generation-1 battle simulator and implemented four heuristic AIs (Random, Attack, Smart-Attack, Smart-Defense) to simulate and balance Pokémon gameplay \citep{panumate2016pokemon}. These simple bots modeled human behavior and were used to study “comfortable” game settings using game-refinement theory. In short, these studies demonstrate that search-based and rule-based methods can yield modest performance in Pokémon battles, but they often struggle with the game’s huge state space and complexity. Karten, Nguyen, and Jin (2025) made PokeChamp, an agent that plays Pokémon battles with a minimax search algorithm enhanced with an LLM (large language model). The LLM performs player action sampling, opponent modeling, and value function estimation, and achieves state-of-the-art LLM performance in the Gen 9 OU format \citep{karten2025pok}.

\subsubsection{Reinforcement Learning Approaches} More recent work has applied deep RL to Pokémon. \citet{huang2019self} trained a deep PPO agent via self-play for singles battles. Their approach requires no handcrafted simulator and succeeds on the nondeterministic, partially-observable Pokémon game, yielding a policy on par with state-of-the-art search agents and competitive with human ladder. \citet{simoes2020competitive} similarly used deep RL on a Pokémon battling simulator (using the Poké-env gym interface) to learn competitive play. \citet{wang2024winning} used DRL with PPO and augmented with parallel MCTS to train an AI agent to play Gen 4 random battles at an expert human level. \citet{pleines2025pokemon} took this further by tackling \emph{Pokémon Red}, an open-world RPG. They wrapped an emulator in an OpenAI Gym and trained a PPO agent to reach Cerulean City. Their experiments revealed that naive reward shaping can be exploited (e.g.\ agents skip intermediate challenges if badges give reward), and they underscore the exploration and multi-tasking challenges in Pokémon games. In summary, DRL (especially policy-gradient methods like PPO) has shown promise for learning Pokémon battle strategies from experience, achieving competitive performance without explicit simulators.

\subsubsection{Team Building and Metagame Analysis} Building a strong team is crucial in Pokémon. \citet{simoes2023adversarial} automated team construction under a metagame framework. They set up an \emph{adversarial} model with two agents: a “team builder” that evolves six-Pokémon teams to maximize win-rate, and a “balancer” that adjusts base stats or move sets to encourage diversity in the meta. The team builders use evolutionary or linear optimization to exploit current metagames, while the balancer nudges them to try underused Pokémon. This adversarial setup iteratively yields balanced meta-teams.

\subsubsection{Other Machine Learning Approaches} Some studies use supervised learning on Pokémon data. For example, \citet{charde2019predicting} trained machine learning models to predict the winner of a Pokémon Showdown battle given the game state. This suggests that supervised classifiers can capture strategic patterns.

\section{\vgcbench}

This section introduces \vgcbench, a comprehensive benchmark built on top of \texttt{poke-env} \citep{poke_env}. \vgcbench{} provides users with (1) infrastructure for multi-agent learning and human-play data collection, (2) a diverse suite of baselines, and (3) robust evaluation tools for performance, generalization, exploitability, and human play.

\subsection{Infrastructure}
\label{sec:ours-infra}
\paragraph{Multi-Agent Environment.} We integrate \texttt{poke-env} with PettingZoo \citep{terry2021pettingzoo} to support parallelized population-based and PSRO-style multi-agent learning \citep{lanctot2017unified} for both players and support the VGC format.
To support more controlled training settings, we introduce two toggles: one to exclude the team preview phase and another to disable mirror matches. Disabling team preview skips that stage from the agent’s perspective, initializing each game with a random team preview decision -- potentially encouraging policy exploration in any team preview situation. Disabling mirror matches prevents an agent from facing the same team configuration during training, which is useful when evaluating a specific matchup between two distinct team configurations. Finally, we provide users with a team scraper that collects high-performing teams in real tournaments from VGCPastes \citep{VGCPastes2025}.

\paragraph{Observation Space.}
The default observation encodes information for each of the 12 Pokémon (6 per player) using a mixture of discrete (e.g., Pokémon type) and continuous (e.g., HP) features. Let $g$ be the global features (e.g., weather), $s$ the side-specific features (e.g., light screen), and $p$ the per-Pokémon features. Each Pokémon is represented as a concatenation of $p$, its side’s $s$, and the global $g$ vector, giving a vector of size $g + s + p$. The overall observation is shaped as $12 \times (g + s + p)$. If frame stacking is used with $n$ frames, the observation becomes $n \times 12 \times (g + s + p)$.

\paragraph{Action Space.}
We represent actions in the VGC format as a joint action space, since each turn in doubles matches requires an action for each of the two active Pokémon. The action space is enumerated in Table \ref{tab:action_mapping}. Each Pokémon has 107 available actions, which captures the full space of switching in benched Pokémon and using moves, where moves involve which move is being picked, the intended target, and whether or not the Pokémon is terastallizing (or using another available gimmick). We also unify the team preview action of the battle with the action space for the rest of the battle by simply modeling team preview as two joint "switch-in" actions in a row, providing two Pokémon per joint action for a total of four Pokémon to make the team preview decision.

\paragraph{Human-Play Data Collector.} To facilitate large-scale imitation learning from human players, we provide a parallel data collection pipeline for scraping and parsing Gen 9 VGC battle replays from Pokémon Showdown. With the current number of available replays from Pokémon Showdown, we are able to amass over 700,000 OTS-enabled VGC games. Due to the neutral perspective of battle logs and its purely transitional messages, reconstructing a trajectory of game states cannot be done perfectly. However, because we filter for only OTS-enabled battle logs, we can derive well-approximated states and actions from the logs by replaying the transitions through our environment. Our log reader also has configurable filters to only read logs of players with ratings above a threshold and only the winner of the battle.

\subsection{Baseline Implementations}
\label{sec:baselines}

\vgcbench{} implements a total of 11 baseline agents: three heuristic agents, an LLM agent, a behavior cloning model (BC), a self-play RL agent (SP), a fictitious play RL agent (FP), and a double oracle RL agent (DO), as well as fine-tuning the agent initialized with the BC policy's parameters using SP, FP, and DO  (BCSP, BCFP, BCDO).

\subsubsection{Trained Baselines}

\paragraph{Behavior Cloning.}
We train a {\bf behavior cloning (BC)} policy \citep{gleave2022imitation} to match the distribution of human actions given a state from the dataset $\mathcal{D}$ collected by the human play data collector, without the use of our rating or winner-based data filters. The parameterized policy $\pi_{\theta}$ is trained to imitate human decision-making by minimizing the negative log-likelihood of the demonstrated actions:
\[
\min_{\theta} \ \mathbbm{E}_{(s, a) \sim \mathcal{D}_{R \geq R_{\text{min}}}}[-\log \pi_{\theta}(a | s)].
\]

\paragraph{Multi-Agent Reinforcement Learning.}
We employ an actor-critic implementation of Proximal Policy Optimization (PPO) \citep{schulman2017ppo} to train all reinforcement learning agents. For agents to approach the desired minimax solution of the game via RL, we implement three multi-agent training paradigms from the unified empirical game-theoretic framework \citep{lanctot2017unified} on top of RL: self-play, fictitious play, and double oracle \citep{mcmahan2003planning}. In \textbf{self-play (SP)}, the agent trains against itself. The \textbf{fictitious play (FP)} variant maintains a pool of the agent's past checkpoints, and the agent learns against a uniform distribution of its past selves, resampling the opponent policies after every experience-gathering period. We take the final response policy to the policy pool as the method's output. \textbf{Double oracle (DO)} differs from FP in that it derives the Nash equilibrium distribution from a maintained empirical payoff matrix between all agents in the pool, and uses that distribution to sample opponents. We solve the Nash distribution by solving a linear programming problem because of the game's two-player, zero-sum, and symmetric-payoff nature. Future work could employ more complex meta-game solvers with the Policy Space Response Oracle  \citep{lanctot2017unified}. We also implement baselines that fine-tune the policy from the BC-trained parameters (\textbf{BCSP}, \textbf{BCFP}, \textbf{BCDO}). Each method is trained with the same number of training teams and interaction timesteps. The hyperparameters for RL used for all PPO methods (details in Table~\ref{tab:rl_config}) were tuned to serve as a reasonable starting point for future work.

\paragraph{Policy and Value Network Architecture.} All the policy networks used across the learning baselines embed each agent's moves, items, and abilities into latent representations, and then use an additional aggregation token and a 3-layer Transformer encoder \citep{vaswani2017attention} to aggregate the information of 12 Pokémon (6 from the agent's team and 6 from the opponent's team) in the field. When frame stacking is enabled, we use a second Transformer encoder along the time axis, apply a positional encoding, and use causal masking to process the historical information and only use the logits from the last frame. The output of the Transformer encoder is then projected to the size of the action space by a linear layer into the logits before the softmax layer. The logits at the invalid actions are masked by $-\infty$. We also handle interdependent action constraints, such as ensuring that both Pokémon do not switch into the same replacement. For reinforcement learning methods, we adopt an actor-critic implementation without parameter sharing, though they share the same network architecture.

\subsubsection{Other Baselines}

\paragraph{LLM Agent.} We also provide a basic \textbf{LLM agent} that uses Meta-Llama-3.1-8B-Instruct \citep{grattafiori2024llama}. Each turn, the battle state is converted into a prompt that is then fed into the agent along with an available action space, instructions on how to construct a valid action, and the required format of its response. If the LLM agent does not respond in the correct format, the agent selects a random action. Note that the LLM agent was not thoroughly researched; we acknowledge that it is possible to make a stronger LLM agent with tool use and other techniques \citep{karten2025pok}. However, this LLM agent baseline provides a reasonable starting point on which research can be conducted.

\paragraph{Heuristic Agents.} From \texttt{poke-env}, we inherit 3 rule-based baselines: \textbf{RandomPlayer}, \textbf{MaxBasePowerPlayer}, and \textbf{SimpleHeuristicsPlayer} \citep{poke_env}. The RandomPlayer plays random moves every turn, the MaxBasePowerPlayer greedily chooses the highest-power move from the available options every turn, and the SimpleHeuristicsPlayer has a hard-coded weighted sum of multiple heuristics to select the best move based on the heuristic scores, which considers information such as current HP of the active Pokémon, type matchup viability, accuracy, power of its moves, etc. We extend SimpleHeuristicsPlayer so that it can also play double battles for the VGC format. 

\subsection{Evaluation Methods}
\label{sec:eval-methods}
\vgcbench{} provides multiple methods of evaluation, including cross-evaluation of agents, Alpha-Rank \citep{omidshafiei2019alpha} for ranking agents based on their cross-evaluation performance, training an exploiter policy against an agent to lower-bound the agent's worst-case lose-rate, and evaluating agents over varying team set sizes with the performance and generalization tests.

Let $\mathcal{C}$ be the set of team configurations used for evaluation and $\mathcal{U}(\mathcal{C})$ be the uniform distribution of those teams. The cross-evaluation of a pair of policies $(\pi_i, \pi_j) \in \Pi \times \Pi$ from the policy pool $\Pi$, where $1 \leq i \leq |\Pi|$, $1 \leq j \leq |\Pi|$, and $i \neq j$, can be defined as
\begin{equation}
\text{crossplay}(\pi_i, \pi_j, \mathcal{C}) = \mathbb{E}_{c_1, c_2 \sim \mathcal{U}(\mathcal{C})}\left[\mathbb{E}_{\tau \sim \Gamma_{c_1, c_2}(\pi_i, \pi_j)}[R(s_T)]\right],
\end{equation}
where $\pi_1$ controls team $c_1$, $\pi_2$ controls team $c_2$, and the distribution of trajectories $\Gamma$ is defined in Equation \ref{eq:traj_dist}. Alpha-Rank \citep{omidshafiei2019alpha} provides a way of transforming this cross-play matrix into an ordered ranking of policies, allowing us to determine a best policy among a pool.

The combination of these evaluation primitives enables insight into agent performance, generalization, and exploitability. 
\begin{enumerate}[leftmargin=*]
    \item \textbf{Performance} can be tested by cross-playing agents with teams that every agent had experience with during training. Ideally, as you train agents with more teams, performance would minimally decrease on an individual team basis against agents trained with fewer teams. We define $C_k$ as the set of teams configurations that policy $\pi_k$ was trained on; the evaluation set of teams $\mathcal{C}_\text{eval}$ is constrained by
    \begin{equation}
    \label{performance-equation}
         \mathcal{C}_\text{eval} = \bigcap_{k=1}^{|\Pi|} \mathcal{C}_{k},
    \end{equation}
    \item \textbf{Generalization} can be tested by cross-playing agents with teams that none of them experienced during training; how much an agent trained on more team configurations surpasses those trained on fewer configurations reflects its generalization ability. We constrain the evaluation set of teams $\mathcal{C_\text{eval}}$ by
    \begin{equation}
        \mathcal{C}_\text{eval} \cap \bigcup_{k=1}^{|\Pi|} \mathcal{C}_k = \emptyset.
    \end{equation}
    \item \textbf{Exploitability} can be tested by approximating a best response policy, $\text{BR}(\pi)$, against the agent to be evaluated via RL training and measuring the highest win rate that the best response can achieve \citep{timbers2020approximate}. Whatever maximum win rate the exploiter agent achieves over a sufficiently large number of training steps serves as our exploitability measurement for our agent. We formally define the exploitability of the agent as
    \begin{equation}
    \label{exploit-equation}
         \text{exp}(\pi, \mathcal{C}_{\text{eval}}) = \text{crossplay}(\text{BR}(\pi), \pi, \mathcal{C}_{\text{eval}}).
    \end{equation}
\end{enumerate}

\section{Experiments}

To encourage research in Pokémon VGC, we gather preliminary results interrogating 2 main research questions. In sections \ref{experiments:perf+gen} and \ref{experiments:exploit}, we investigate how performance, generalization, and exploitability scale as the training team size increases. In section \ref{experiments:human-eval} we see if it is possible to use standard RL methods to create an AI agent that plays at a human expert level for the one-team setting. All experiments are conducted on a cluster with 8 A40 GPUs and 2 Intel(R) Xeon(R) Gold 6342 CPU @ 2.80GHz.

\subsection{Cross-Play Performance}

\begin{figure*}[h!]
    \centering
    \includegraphics[width=\linewidth]{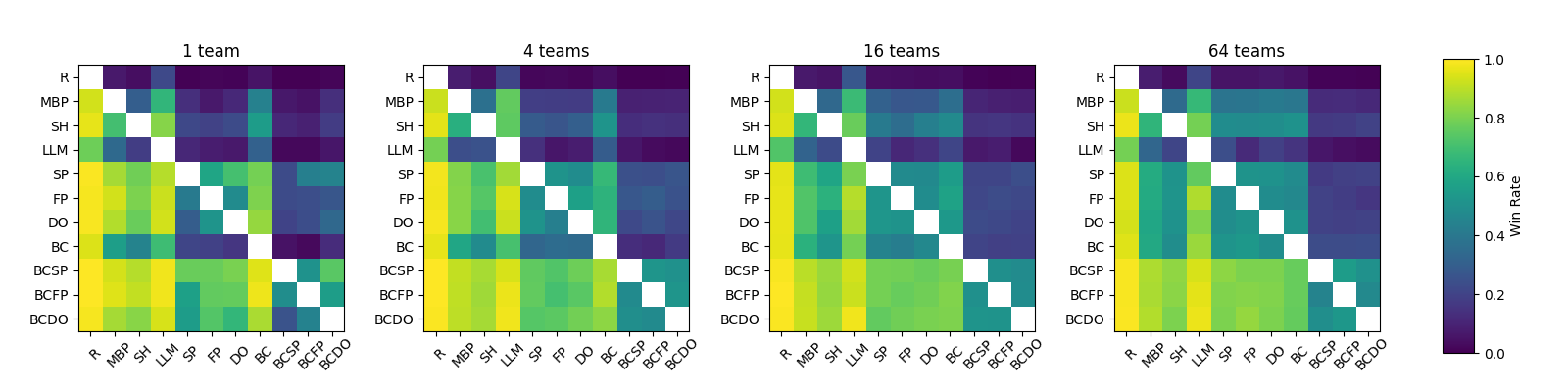}
    \caption[Cross-Play Win Rate Heatmaps]{\small \textit{\textbf{Cross-Play Win Rate Heatmaps}} for varying team set sizes.}
    \label{fig:heatmaps}
    \Description{Cross-Play Win Rate Heatmaps for varying team set sizes.}
\end{figure*}

For all agents that fine-tuned with RL, we did training runs with team set sizes of 1, 4, 16, and 64. For each team set size, we compare our 6 RL baseline agents, the BC agent, 3 rule-based players, and our LLM agent (all described in section \ref{sec:baselines}) through cross-play in Figure~\ref{fig:heatmaps}. Each win-rate entry of the cross-play matrix is calculated with 1000 episodes with uniformly drawn teams for each game played, except for the LLM player, which was evaluated with 100 episodes due to the LLM's relatively slow inference. These 1000 episodes equally evaluate across 5 training seeds for all learning methods. For each seed, a different set of teams is selected for the 1, 4, 16, and 64-team settings, and these team sets are always perfectly nested so that, for example, all teams from the 16-team setting of a given seed are included in the 64-team setting of that seed. All cross-evaluations have win-rates relative to the row players, with errors of approximately $\pm 0.03$ for 1000-game evaluations, and $\pm 0.10$ for the LLM player's 100-game evaluations. For numerical win rates, see Tables \ref{tab:ai_comparison_1_team}, \ref{tab:ai_comparison_4_teams}, \ref{tab:ai_comparison_16_teams}, \ref{tab:ai_comparison_64_teams} in Appendix \ref{sec:additional-tables}.

Notice that as the number of teams increases, the heatmaps become progressively less noisy. We hypothesize that this is likely caused by (1) all of our learning methods noticeably degrading in performance for larger team set sizes, and (2) results being averaged over larger numbers of matchups reducing matchup-specific biases in the evaluation.

\begin{table}[h!]
\centering
\caption[Alpha-Rank Results]{\small \textit{\textbf{Alpha-Rank Results.}} This table contains a relative ranking of each baseline agent featured in \vgcbench. Each agent is only compared with other agents that trained with the same number of teams.}
\setlength{\tabcolsep}{3pt}
\begin{tabular}{lcccc}
\toprule
 & \textbf{1 Team} & \textbf{4 Teams} & \textbf{16 Teams} & \textbf{64 Teams} \\
\midrule
R     & 11 & 9 & 11 & 11 \\
MBP   & 9  & 7 & 9  & 9  \\
SH    & 7  & 5 & 8  & 7  \\
LLM   & 10 & 8 & 10 & 10 \\
SP    & 4  & 4 & 6  & 4  \\
FP    & 6  & 4 & 5  & 8  \\
DO    & 5  & 4 & 4  & 5  \\
BC    & 8  & 6 & 7  & 6  \\
BCSP  & \textbf{1} & \textbf{1} & 3 & \textbf{1} \\
BCFP  & 2  & 2 & 2  & 3  \\
BCDO  & 3  & 3 & \textbf{1} & 2  \\
\bottomrule
\end{tabular}
\label{tab:alpharank}
\Description{This table contains a relative ranking of each baseline agent featured in \vgcbench. Each agent is only compared with other agents that trained with the same number of teams.}
\end{table}

It is clear from Figure \ref{fig:heatmaps} and Table \ref{tab:alpharank} that while non-BC methods achieve decent performance, they are generally outperformed by methods initialized with BC. Self play seems to be the most reliably high-performance method in the 1 team setting, but we note that the performance differences between the BC methods for the 4, 16, and 64 team set sizes are negligible.

\subsection{Performance and Generalization Evaluations}
\label{experiments:perf+gen}
We then measure performance and generalization as described in section \ref{sec:eval-methods}. We used the highest overall ranked agents from our Alpha-Rank evaluation in the 1, 4, 16, and 64-team settings for this evaluation; Table \ref{tab:performance_test} shows that as the size of the set of teams used during training increases, the performance of the AI agent on any one team decreases considerably and consistently. These results indicate that our agent does \textbf{not} pass the performance test even for relatively small $n \leq 64$, highlighting the central challenge that Pokémon VGC poses to the AI community. To test generalization, we used 72 out-of-distribution teams and compared agents across team set sizes via cross-play. In Table \ref{tab:generalization_test}, we can see that our agent exhibits generalization since win rate increases moderately as $n$ increases.

\begin{table}[h!]
\centering
\caption[Performance Test]{\small \textit{\textbf{Performance Test.}} Each matchup evaluated for 1000 games, with errors $\pm 0.03$.}
\setlength{\tabcolsep}{4pt}
\begin{tabular}{lcccc}
\toprule
\textbf{\#Teams} & \textbf{1 (BCSP)} & \textbf{4 (BCSP)} & \textbf{16 (BCDO)} & \textbf{64 (BCSP)} \\
\midrule
\textbf{1 (BCSP)} & -- & 0.699 & 0.74 & 0.698 \\
\textbf{4 (BCSP} & 0.301 & -- & 0.594 & 0.672 \\
\textbf{16 (BCDO)} & 0.26 & 0.406 & -- & 0.644 \\
\textbf{64 (BCSP)} & 0.302 & 0.328 & 0.356 & -- \\
\bottomrule
\end{tabular}
\label{tab:performance_test}
\end{table}

\hfill

\begin{table}[h!]
\centering
\caption[Generalization Test]{\small \textit{\textbf{Generalization Test.}} Each matchup evaluated for 1000 games, with errors $\pm 0.03$.}
\setlength{\tabcolsep}{4pt}
\begin{tabular}{lcccc}
\toprule
\textbf{\#Teams} & \textbf{1 (BCSP)} & \textbf{4 (BCSP)} & \textbf{16 (BCDO)} & \textbf{64 (BCSP)} \\
\midrule
\textbf{1 (BCSP} & -- & 0.405 & 0.375 & 0.331 \\
\textbf{4 (BCSP)} & 0.595 & -- & 0.453 & 0.422 \\
\textbf{16 (BCDO)} & 0.625 & 0.547 & -- & 0.436 \\
\textbf{64 (BCSP)} & 0.669 & 0.578 & 0.564 & -- \\
\bottomrule
\end{tabular}
\label{tab:generalization_test}
\end{table}

In order to solidify the validity of our generalization results, we also calculate statistics regarding the similarity of the 72 teams we set aside during training with the upper bound of 64 teams used during training in Table \ref{tab:team_similarity}. We measure team similarity on a scale of 0 to 1, taking into account matching species between teams, and among those matching species, whether they have the same item, moves, ability, Tera type, and stat configurations. We ensure that no two teams are fully identical, and assert that the average similarity being about 0.5 across all training seeds firmly grounds the Generalization Test as a legitimate experiment.

\begin{table}[h]
\centering
\caption[Team Similarity Scores]{\small \textit{\textbf{Team Similarity Score Statistics.}} Comparison between the $\leq 64$ seen teams and the 72 unseen teams.}
\begin{tabular}{lcccc}
\toprule
\textbf{Run} & \textbf{Mean} & \textbf{Median} & \textbf{Min} & \textbf{Max} \\
\midrule
1 & 0.508 & 0.513 & 0.268 & 0.946 \\
2 & 0.522 & 0.526 & 0.221 & 0.947 \\
3 & 0.530 & 0.524 & 0.221 & 0.948 \\
4 & 0.557 & 0.554 & 0.260 & 0.947 \\
5 & 0.526 & 0.526 & 0.238 & 0.948 \\
\bottomrule
\end{tabular}
\label{tab:team_similarity}
\end{table}

\subsection{Exploitability Evaluation}
\label{experiments:exploit}
\begin{figure}[h]
    \centering
    \includegraphics[width=1\linewidth]{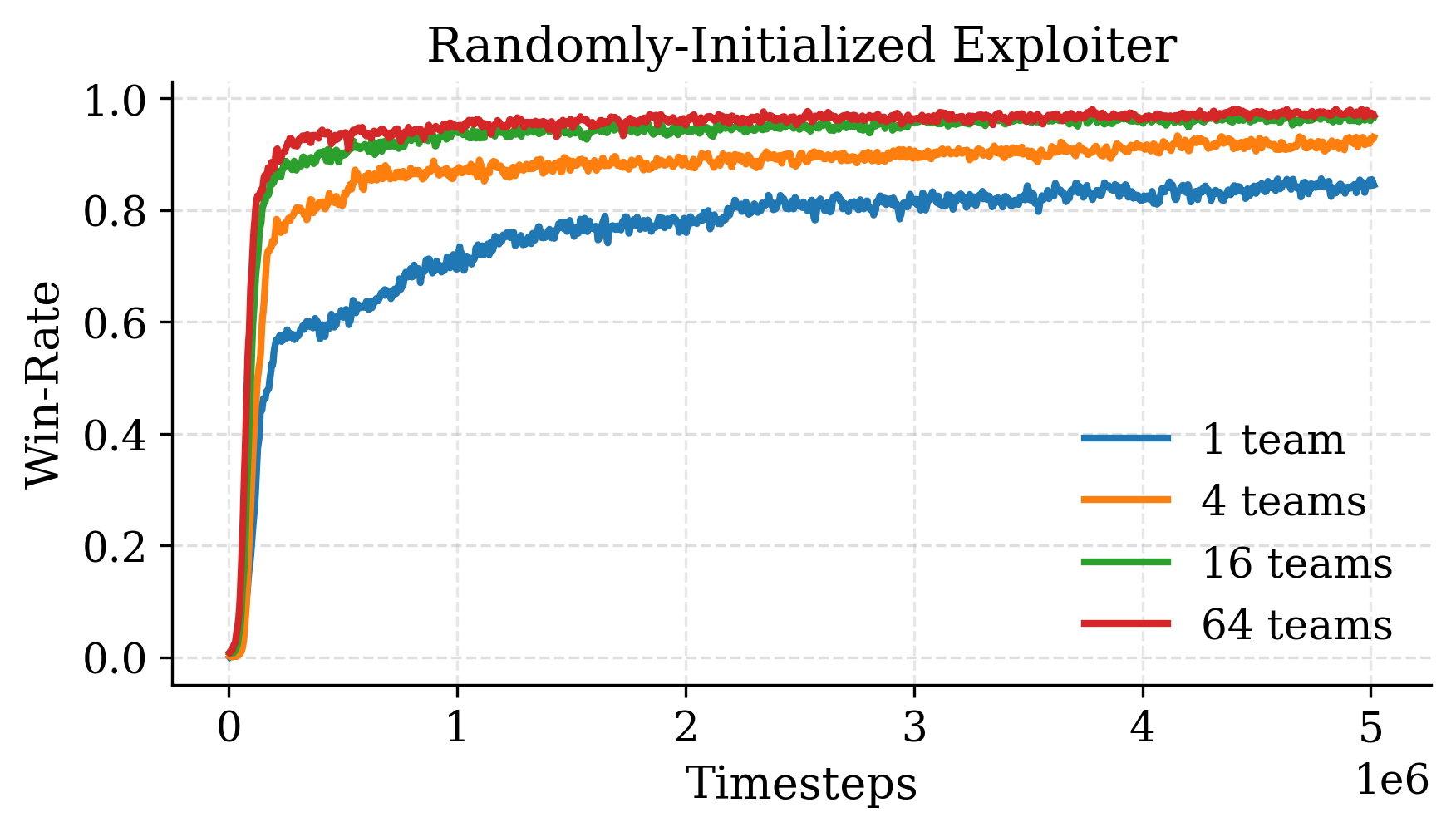}
    \caption[Randomly-initialized Exploiter]{\small An exploiter agent initialized randomly trains to exploit the strongest agent as determined in Table \ref{tab:alpharank}.}
    \label{fig:exploit}
    \Description{An exploiter agent initialized randomly trains to exploit the strongest agent as determined in Table \ref{tab:alpharank}.}
\end{figure}

\begin{figure}[h]
    \centering
    \includegraphics[width=1\linewidth]{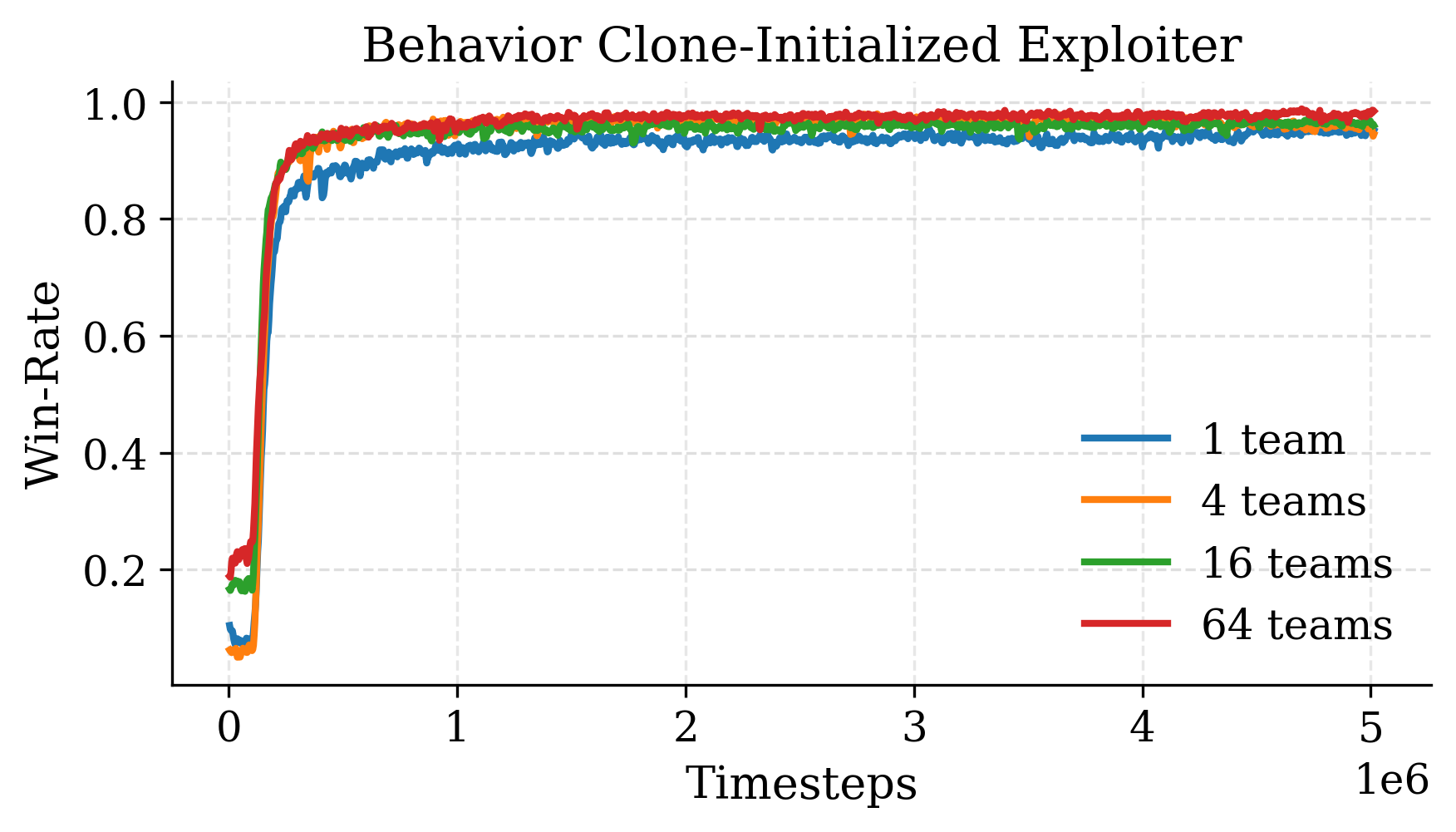}
    \caption[Behavior Clone-Initialized Exploiter]{\small An exploiter agent initialized as the behavior cloning agent trains to exploit the strongest agent as determined in Table \ref{tab:alpharank}.}
    \label{fig:exploit-bc}
    \Description{An exploiter agent initialized as the behavior cloning agent trains to exploit the strongest agent as determined in Table \ref{tab:alpharank}.}
\end{figure}

We measured the exploitability of the strongest agent according to the Alpha-Rank evaluation from each of our team set sizes as described in section \ref{sec:eval-methods}. This evaluation was averaged across the agents from all 5 training seeds, and all exploiter agents were trained on the same team configuration from the 1-team setting, regardless of how many teams the agent being exploited experienced during training. We tested exploitability by initializing the exploiter with a random initialization in Table \ref{fig:exploit} and with the pre-trained BC policy in Table \ref{fig:exploit-bc}. In almost all cases, all agents are approximately 100\% exploitable, although the best agent in the 1-team setting does exhibit notably stronger resistance to exploitation when the exploiter agent is randomly initialized.

\subsection{Human Evaluation}
\label{experiments:human-eval}
After measuring Alpha-Rank ratings and evaluating performance across team set sizes, we tested our most performant agent in the most performant team set size -- the behavior cloning agent fine-tuned with self-play in the 1-team setting -- against an intermediate, advanced, and expert player. Our expert player was Aaron Traylor, a multi-time competitor in the World Championships and the 2020 Dallas Regional Champion. All 3 human testers were instructed to try to find a way to exploit the AI system over 5 consecutive games and encouraged to think for as long as they needed to. Against the intermediate player, our agent won all 5 matches, and against the advanced player, the agent won 2 out of 5 total matches played. Indeed, our agent was even able to win against the expert-level player. To the best of our knowledge, our agent is the first AI system to achieve such a feat. We received feedback that although the agent is strong on initial play, it does have noticeable dips in performance in certain states. After enough successive games, strong human players can adapt and beat the agent. Please note that these human evaluations are intentionally anecdotal; more comprehensive human evaluation should be conducted before reaching stronger conclusions based on these results.

\section{Conclusion and Future Work}

In this work, we introduced \vgcbench, a benchmark designed to evaluate the generalization capabilities of AI agents in the challenging and combinatorially complex environment of Pokémon VGC. Our benchmark includes standardized evaluation protocols, a modular training pipeline, curated human gameplay data, and implementations of a broad set of baseline methods ranging from behavior cloning and reinforcement learning with game-theoretic algorithms like self-play, fictitious play, and double oracle, the strongest of which was able to win against a past World Championships competitor. We also contributed substantial improvements to the Poké-env library, including integration with PettingZoo and environment support for VGC formats, thereby enabling easy adoption by the broader research community.

Through extensive experiments, we demonstrated that while current algorithms can attain strong performance in the single-team setting, they degrade significantly when scaling to multi-team generalization -- highlighting a key open challenge in multi-agent learning. By providing a reproducible benchmark and uncovering a difficult generalization frontier, our work establishes a foundation for future progress in robust multi-agent policy learning.

Looking forward, we identify five major research directions enabled by \vgcbench:

\begin{enumerate}[leftmargin=*]
\item \textbf{Generalization to $n$ Teams, $n > 1$:} Our current agents achieve strong performance in the single-team setting but struggle as the number of teams in training increases. A natural extension is to develop agents that can generalize across team matchups without having performance degrade, and ultimately perform at a superhuman level across arbitrary teams without needing to retrain. Researchers can use the experiments in Tables~\ref{tab:performance_test} and~\ref{tab:generalization_test} to assess progress on this front.
\item \textbf{Team Building:} A strong VGC agent could be used to evaluate candidate teams and provide a reward signal for searching the vast team configuration space. While such work would be downstream of generalization, our current agents may already offer a useful starting point for this line of research.
\item \textbf{Search:} With an accurate and fast simulator, one could model future positions of the game to make more well-informed decisions \citep{brown2020combining, sokota2025superhuman}. The benefits of model-based techniques could be bountiful in the domain of Pokémon VGC, as it gives the agent a way to understand how the individual actions of each active Pokémon on the field contribute to the resulting state. In fact, we suspect that much of the difficulty of achieving strong performance across a large set of teams with model-free RL may be that the complexity of predicting future outcomes of the game simply can't fit into the deep neural network. Past work in Pokémon suggests that search has significant potential \citep{huang2019self, wang2024winning, karten2025pok}, but active work is ongoing to create a highly accurate, fast, and open-source simulator.
\item \textbf{Opponent Modeling:} One could incorporate explicit opponent modeling to infer latent properties of the opposing strategy, such as playstyle, risk tolerance, and long-term tactical intent. By predicting how an opponent is likely to act -- or which strategic archetype they belong to -- an agent can dynamically adapt its policy rather than treating all opponents as identically optimal or uniformly random \citep{he2016opponent, yu2022model}. It may be a good idea to start the opponent model out as a behavior cloning agent from human data and attempt to converge to the opponent's style as the game plays out, effectively adapting to the opponent in real time from a reasonable starting guess.
\item \textbf{Latent Variable Behavior Cloning:} Future work could use latent variable imitation learning, enabling the behavior cloning agent to model diverse human play-styles rather than collapsing behavior into a single averaged policy, which could improve its strength by making its play more cohesive and consistent \citep{wang2017robust, hsiao2019learning}.
\end{enumerate}



\begin{acks}
\sloppy
This work has taken place in the Learning Agents Research
Group (LARG) at UT Austin.  LARG research is supported in part by NSF
(FAIN-2019844, NRT-2125858), ONR (W911NF-25-1-0065), ARO
(W911NF-23-2-0004), DARPA (Cooperative Agreement HR00112520004 on Ad
Hoc Teamwork) Lockheed Martin, and UT Austin's Good Systems grand
challenge.  Peter Stone serves as the Chief Scientist of Sony AI and
receives financial compensation for that role.  The terms of this
arrangement have been reviewed and approved by the University of Texas
at Austin in accordance with its policy on objectivity in research.

We are very thankful to our human testers: Dylan Remillard (intermediate), Jiaheng Hu (advanced), and Aaron Traylor (expert). Aaron Traylor was also kind enough to explain some niche mechanics in Pokémon VGC, give helpful feedback to the AI's strengths and weaknesses, connect us with other professional VGC competitors, and even used their platform to help popularize our work. Also, we thank our anonymous reviewers for their diligence in evaluating our work and for the constructive criticism which ultimately helped this paper become what it is.
\end{acks}



\bibliographystyle{ACM-Reference-Format}
\balance
\bibliography{refs}


\clearpage
\appendix

\section{Game Analysis}
\label{sec:game-analysis}
Pokémon VGC is a complex game; there are up to four Pokémon out at a time, each able to target up to three targets with a move and potentially terastallize or switch out into up to two benched Pokémon. Then we can calculate the approximate branching factor of nondeterminism in Pokémon moves. Calculated loosely, there are 16 damage rolls for each damaging move, almost always with a chance to hit normally, fail, land a critical hit, land an additional effect, or both land a critical hit and an additional effect. If we only consider these possibilities, there are $16 * 5 = 80$ outcomes for every move used. Also, each Pokémon has 4 total moves they can use, each of which have a maximum of 3 possible targets. If terastallization is available, there are
\[
2 \cdot 80 \cdot \mathrm{n\_moves} \cdot \mathrm{n\_targets} + \mathrm{n\_switches} = 2 \cdot 80 \cdot 3 \cdot 4 + 2 = 1922
\]
possible outcomes, and without terastallization there are
\[
80 \cdot \mathrm{n\_moves} \cdot \mathrm{n\_targets} + \mathrm{n\_switches} = 80 \cdot 3 \cdot 4 + 2 = 962
\]
possible outcomes. Thus, in a turn, there is a worst-case branching factor of approximately
\[
1922^2 \cdot 962^2 = 3.419 \times 10^{12} \approx 10^{12}.
\]

Additionally, we can quantify the worst-case size of a player's information set, or the number of indistinguishable states due to Pokémon VGC's partial observability. The only unobservables are the opponent Pokémon's stats and the 2 bench Pokémon the opponent selected during team preview. For the latter, there are $\binom{4}{2} = 6$ possibilities. As for stats, Pokémon stats are determined by a Pokémon's EV's, IV's, and natures. There are 25 natures (although 5 are identical in effect, so we will use 21), 32 possible values for each of its 6 IV values, and 512 points to distribute over 6 stats for EV values, with no more than 252 points for each stat. Assuming all EV points are used and all EV spreads would distribute points to stats that are divisible by 4, which is reasonable because EV points only affect stats in multiples of 4, we can formalize EV spreading as the problem of distributing $510 // 4 = 127$ units across 6 stats, with the maximum allocation per stat being $255 // 4 = 63$. We must find the total number of solutions to the equation $x_1 + x_2 + x_3 + x_4 + x_5 + x_6 = 127$, $0 \leq x_i \leq 63$ for $1 \leq i \leq 6$. This can be solved using the inclusion-exclusion principle for bounded integer compositions, and yields a total count of 246,774,528. We will leave IVs out of the calculation since most Pokémon are given all 31 points per stat. Thus, the number of possible realistic stat configurations is 
\[\left(21 \cdot 246774528\right)^{\mathrm{n\_mons}} = 1.937 \times 10^{58} \approx 10^{58}.
\]
This is the best-case size of the information set if the opponent has revealed all four Pokémon they chose during team preview; if they haven't, then the worst-case size of the information set is
\[
6 \cdot 1.937 \times 10^{58} = 1.162 \times 10^{59} \approx 10^{59}.
\]

Let us now attempt a rough count of the total number of possible team configurations in VGC in Pokémon Scarlet and Violet. With 937 total moves in the game (of which we will approximate about 100 are available in a given Pokémon's learnset), up to 3 available abilities for a species, 540 holdable items (of which only 223 are available in the latest game and are not pokeballs), 2 genders (usually), as well as the choice of a tera type and stat configurations for each Pokémon (calculated above to be $21 * 246774528 = 5.182 \times 10^9$), the total number of configurations for a single Pokémon can be calculated as

\[
    \binom{100}{4} \cdot 3 \cdot 223 \cdot 2 \cdot 19 \cdot 5.182 \times 10^9 = 5.166 \times 10^{20}.
\]

Thus, with approximately 750 species available in the current games as of the writing of this paper, the total number of team configurations is

\[
    \binom{750}{6} \cdot (5.166 \times 10^{20})^6 = 4.604 \times 10^{138} \approx 10^{139}.
\]

In chess and Go, players can only initialize as white or black. In Texas hold 'em Poker, each player is dealt 2 cards out of a 52-card deck at the beginning of the game, yielding a total initialization space of $\prod_{i=0}^{n-1}\binom{52-2i}{2}$. Dota has two teams of five players choose from a total of 126 available heroes, yielding a total configuration space of $\binom{126}{5}\binom{121}{5} = 4.85 \times 10^{16} \approx 10^{17}$. Finally, Starcraft allows each of the two players to choose from three available races, as well as 9 playable maps, yielding $9 * 3^2 = 81$ total configurations.

\section{Additional Tables}
\label{sec:additional-tables}

\begin{table}[H]
\centering
\caption[Action Space]{\small \textit{\textbf{Per-Slot Action Space in Doubles Format.}} Moves have ranges of length 5 due to all possible targets: slot a self, slot b self, no target, slot a opponent, slot b opponent.}
\begin{tabular}{|c|l|}
\hline
\textbf{Index} & \textbf{Action Description} \\
\hline
-2 & Default \\
-1 & Forfeit \\
0 & Pass \\
1--6 & Switch \\
7--11 & Move 1 \\
12--16 & Move 2 \\
17--21 & Move 3 \\
22--26 & Move 4 \\
27--31 & Move 1 + Mega Evolve \\
32--36 & Move 2 + Mega Evolve \\
37--41 & Move 3 + Mega Evolve \\
42--46 & Move 4 + Mega Evolve \\
47--51 & Move 1 + Z-Move \\
52--56 & Move 2 + Z-Move \\
57--61 & Move 3 + Z-Move \\
62--66 & Move 4 + Z-Move \\
67--71 & Move 1 + Dynamax \\
72--76 & Move 2 + Dynamax \\
77--81 & Move 3 + Dynamax \\
82--86 & Move 4 + Dynamax \\
87--91 & Move 1 + Terastallize \\
92--96 & Move 2 + Terastallize \\
97--101 & Move 3 + Terastallize \\
102--106 & Move 4 + Terastallize \\
\hline
\end{tabular}
\label{tab:action_mapping}
\end{table}

\begin{table}[H]
\centering
\caption[RL Configuration]{\small \textit{\textbf{Reinforcement Learning Experiment Configuration.}}}
\label{tab:rl_config}
\begin{tabular}{ll}
\hline
\textbf{Hyperparameter} & \textbf{Value} \\
\hline
\textbf{Learning Rate} & 1e-5 \\
\textbf{Discount Factor ($\gamma$)} & 1.0 \\
\textbf{GAE Lambda ($\lambda$)} & 0.95 \\
\textbf{Clip Range} & 0.2 \\
\textbf{Entropy Coefficient} & 0.001 \\
\textbf{Value Function Coefficient} & 0.5 \\
\textbf{Max Gradient Norm} & 0.5 \\
\textbf{Number of Steps per Update} & 24 * 128 \\
\textbf{Batch Size} & 64 \\
\textbf{Number of Epochs} & 10 \\
\textbf{Total Timesteps} & 5,013,504 \\
\hline
\end{tabular}
\end{table}

\begin{table}[H]
\centering
\caption{\small \textit{\textbf{Cross-evaluation Win Rate.}} 1 team, 1000 games, 100 games for LLM player, win-rate relative to row player, errors $\pm 0.03$.}
\begin{adjustbox}{width=\columnwidth,center}
\begin{tabular}{lccccccccccc}
\toprule
\textbf{Algorithm} & \textbf{R} & \textbf{MBP} & \textbf{SH} & \textbf{LLM} & \textbf{SP} & \textbf{FP} & \textbf{DO} & \textbf{BC} & \textbf{BCSP} & \textbf{BCFP} & \textbf{BCDO} \\
\midrule
\textbf{R} & -- & 0.068 & 0.039 & 0.220 & 0.007 & 0.014 & 0.008 & 0.053 & 0.001 & 0.002 & 0.012 \\
\textbf{MBP} & 0.932 & -- & 0.298 & 0.660 & 0.134 & 0.070 & 0.116 & 0.440 & 0.064 & 0.050 & 0.136 \\
\textbf{SH} & 0.961 & 0.702 & -- & 0.820 & 0.215 & 0.196 & 0.229 & 0.551 & 0.110 & 0.091 & 0.178 \\
\textbf{LLM} & 0.780 & 0.340 & 0.180 & -- & 0.110 & 0.080 & 0.070 & 0.310 & 0.020 & 0.020 & 0.060 \\
\textbf{SP} & 0.993 & 0.866 & 0.785 & 0.890 & -- & 0.587 & 0.707 & 0.790 & 0.229 & 0.432 & 0.448 \\
\textbf{FP} & 0.986 & 0.930 & 0.804 & 0.920 & 0.413 & -- & 0.482 & 0.805 & 0.229 & 0.239 & 0.266 \\
\textbf{DO} & 0.992 & 0.884 & 0.771 & 0.930 & 0.293 & 0.518 & -- & 0.842 & 0.203 & 0.236 & 0.337 \\
\textbf{BC} & 0.947 & 0.560 & 0.449 & 0.690 & 0.210 & 0.195 & 0.158 & -- & 0.048 & 0.025 & 0.126 \\
\textbf{BCSP} & 0.999 & 0.936 & 0.890 & 0.980 & 0.771 & 0.771 & 0.797 & 0.952 & -- & 0.510 & 0.744 \\
\textbf{BCFP} & 0.998 & 0.950 & 0.909 & 0.980 & 0.568 & 0.761 & 0.764 & 0.975 & 0.490 & -- & 0.556 \\
\textbf{BCDO} & 0.988 & 0.864 & 0.822 & 0.940 & 0.552 & 0.734 & 0.663 & 0.874 & 0.256 & 0.444 & -- \\
\bottomrule
\end{tabular}
\end{adjustbox}
\label{tab:ai_comparison_1_team}
\end{table}

\begin{table}[H]
\centering
\caption{\small \textit{\textbf{Cross-evaluation Win Rate.}} 4 teams, 1000 games, 100 games for LLM player, win-rate relative to row player, errors $\pm 0.03$.}
\begin{adjustbox}{width=\columnwidth,center}
\begin{tabular}{lccccccccccc}
\toprule
\textbf{Algorithm} & \textbf{R} & \textbf{MBP} & \textbf{SH} & \textbf{LLM} & \textbf{SP} & \textbf{FP} & \textbf{DO} & \textbf{BC} & \textbf{BCSP} & \textbf{BCFP} & \textbf{BCDO} \\
\midrule
\textbf{R} & -- & 0.079 & 0.041 & 0.210 & 0.016 & 0.023 & 0.013 & 0.036 & 0.003 & 0.002 & 0.007 \\
\textbf{MBP} & 0.921 & -- & 0.369 & 0.760 & 0.187 & 0.182 & 0.176 & 0.407 & 0.091 & 0.094 & 0.098 \\
\textbf{SH} & 0.959 & 0.631 & -- & 0.750 & 0.289 & 0.262 & 0.302 & 0.518 & 0.132 & 0.143 & 0.133 \\
\textbf{LLM} & 0.790 & 0.240 & 0.250 & -- & 0.140 & 0.060 & 0.080 & 0.290 & 0.060 & 0.030 & 0.020 \\
\textbf{SP} & 0.984 & 0.813 & 0.711 & 0.860 & -- & 0.513 & 0.488 & 0.671 & 0.246 & 0.239 & 0.263 \\
\textbf{FP} & 0.977 & 0.818 & 0.738 & 0.940 & 0.487 & -- & 0.569 & 0.647 & 0.273 & 0.293 & 0.253 \\
\textbf{DO} & 0.987 & 0.824 & 0.698 & 0.920 & 0.512 & 0.431 & -- & 0.653 & 0.222 & 0.256 & 0.213 \\
\textbf{BC} & 0.964 & 0.593 & 0.482 & 0.710 & 0.329 & 0.353 & 0.347 & -- & 0.130 & 0.114 & 0.171 \\
\textbf{BCSP} & 0.997 & 0.909 & 0.868 & 0.940 & 0.754 & 0.727 & 0.778 & 0.870 & -- & 0.523 & 0.505 \\
\textbf{BCFP} & 0.998 & 0.906 & 0.857 & 0.970 & 0.761 & 0.707 & 0.744 & 0.886 & 0.477 & -- & 0.522 \\
\textbf{BCDO} & 0.993 & 0.902 & 0.867 & 0.980 & 0.737 & 0.747 & 0.787 & 0.829 & 0.495 & 0.478 & -- \\
\bottomrule
\end{tabular}
\end{adjustbox}
\label{tab:ai_comparison_4_teams}
\end{table}

\begin{table}[H]
\centering
\caption{\small \textit{\textbf{Cross-evaluation Win Rate.}} 16 teams, 1000 games, 100 games for LLM player, win-rate relative to row player, errors $\pm 0.03$.}
\begin{adjustbox}{width=\columnwidth,center}
\begin{tabular}{lccccccccccc}
\toprule
\textbf{Algorithm} & \textbf{R} & \textbf{MBP} & \textbf{SH} & \textbf{LLM} & \textbf{SP} & \textbf{FP} & \textbf{DO} & \textbf{BC} & \textbf{BCSP} & \textbf{BCFP} & \textbf{BCDO} \\
\midrule
\textbf{R} & -- & 0.067 & 0.052 & 0.270 & 0.040 & 0.037 & 0.032 & 0.038 & 0.010 & 0.003 & 0.005 \\
\textbf{MBP} & 0.933 & -- & 0.337 & 0.680 & 0.310 & 0.272 & 0.274 & 0.361 & 0.104 & 0.089 & 0.084 \\
\textbf{SH} & 0.948 & 0.663 & -- & 0.770 & 0.413 & 0.358 & 0.432 & 0.478 & 0.152 & 0.160 & 0.147 \\
\textbf{LLM} & 0.730 & 0.320 & 0.230 & -- & 0.200 & 0.110 & 0.140 & 0.210 & 0.070 & 0.080 & 0.020 \\
\textbf{SP} & 0.960 & 0.690 & 0.587 & 0.800 & -- & 0.479 & 0.474 & 0.551 & 0.209 & 0.208 & 0.241 \\
\textbf{FP} & 0.963 & 0.728 & 0.642 & 0.890 & 0.521 & -- & 0.486 & 0.578 & 0.211 & 0.232 & 0.216 \\
\textbf{DO} & 0.968 & 0.726 & 0.568 & 0.860 & 0.526 & 0.514 & -- & 0.533 & 0.228 & 0.215 & 0.200 \\
\textbf{BC} & 0.962 & 0.639 & 0.522 & 0.790 & 0.449 & 0.422 & 0.467 & -- & 0.204 & 0.191 & 0.193 \\
\textbf{BCSP} & 0.990 & 0.896 & 0.848 & 0.930 & 0.791 & 0.789 & 0.772 & 0.796 & -- & 0.498 & 0.481 \\
\textbf{BCFP} & 0.997 & 0.911 & 0.840 & 0.920 & 0.792 & 0.768 & 0.785 & 0.809 & 0.502 & -- & 0.486 \\
\textbf{BCDO} & 0.995 & 0.916 & 0.853 & 0.980 & 0.759 & 0.784 & 0.800 & 0.807 & 0.519 & 0.514 & -- \\
\bottomrule
\end{tabular}
\end{adjustbox}
\label{tab:ai_comparison_16_teams}
\end{table}

\begin{table}[H]
\centering
\caption{\small \textit{\textbf{Cross-evaluation Win Rate.}} 64 teams, 1000 games, 100 games for LLM player, win-rate relative to row player, error $\pm 0.03$.}
\begin{adjustbox}{width=\columnwidth,center}
\begin{tabular}{lccccccccccc}
\toprule
\textbf{Algorithm} & \textbf{R} & \textbf{MBP} & \textbf{SH} & \textbf{LLM} & \textbf{SP} & \textbf{FP} & \textbf{DO} & \textbf{BC} & \textbf{BCSP} & \textbf{BCFP} & \textbf{BCDO} \\
\midrule
\textbf{R} & -- & 0.081 & 0.029 & 0.210 & 0.054 & 0.053 & 0.066 & 0.050 & 0.008 & 0.010 & 0.006 \\
\textbf{MBP} & 0.919 & -- & 0.344 & 0.670 & 0.386 & 0.388 & 0.407 & 0.398 & 0.122 & 0.127 & 0.117 \\
\textbf{SH} & 0.971 & 0.656 & -- & 0.790 & 0.486 & 0.482 & 0.490 & 0.511 & 0.166 & 0.173 & 0.199 \\
\textbf{LLM} & 0.790 & 0.330 & 0.210 & -- & 0.240 & 0.120 & 0.190 & 0.150 & 0.060 & 0.040 & 0.030 \\
\textbf{SP} & 0.946 & 0.614 & 0.514 & 0.760 & -- & 0.512 & 0.510 & 0.481 & 0.169 & 0.189 & 0.196 \\
\textbf{FP} & 0.947 & 0.612 & 0.518 & 0.880 & 0.488 & -- & 0.487 & 0.466 & 0.198 & 0.180 & 0.160 \\
\textbf{DO} & 0.934 & 0.593 & 0.510 & 0.810 & 0.490 & 0.513 & -- & 0.508 & 0.197 & 0.189 & 0.198 \\
\textbf{BC} & 0.950 & 0.602 & 0.489 & 0.850 & 0.519 & 0.534 & 0.492 & -- & 0.231 & 0.234 & 0.238 \\
\textbf{BCSP} & 0.992 & 0.878 & 0.834 & 0.940 & 0.831 & 0.802 & 0.803 & 0.769 & -- & 0.554 & 0.505 \\
\textbf{BCFP} & 0.990 & 0.873 & 0.827 & 0.960 & 0.811 & 0.820 & 0.811 & 0.766 & 0.446 & -- & 0.469 \\
\textbf{BCDO} & 0.994 & 0.883 & 0.801 & 0.970 & 0.804 & 0.840 & 0.802 & 0.762 & 0.495 & 0.531 & -- \\
\bottomrule
\end{tabular}
\end{adjustbox}
\label{tab:ai_comparison_64_teams}
\end{table}


\end{document}